%% file: main.tex
\documentclass[11pt]{article}
\usepackage[
    top=1in,
    bottom=1in,
    left=0.6in,
    right=0.6in
]{geometry}
\usepackage{graphicx}
\usepackage{xcolor}
\usepackage{booktabs}
\usepackage{longtable}
\usepackage{ltablex}
\keepXColumns
\usepackage{array}
\usepackage{titling}
\usepackage{multirow}
\usepackage{tabularx} 
\usepackage{longtable}
\usepackage{float}
\usepackage[skip=5pt]{caption} 
\usepackage{enumitem}
\usepackage{url}
\usepackage[
  colorlinks=true,
  linkcolor=blue,
  citecolor=blue,
  urlcolor=blue
]{hyperref}

\usepackage{cite}

\makeatletter
\let\cite@orig\cite
\renewcommand{\cite}[1]{\textsuperscript{\cite@orig{#1}}}
\renewcommand\@biblabel[1]{#1.}
\makeatother

\title{Vision-centric generative AI models:\\
 A software-hardware perspective
}

\author{
    Eleni Tselepi,
    Cristian Sestito,\\
    Shady Agwa, 
    Themis Prodromakis
}

\date{
    \small
    Centre for Electronics Frontiers, Institute for Integrated Micro and Nano Systems, \\ School of Engineering, The University of Edinburgh, UK
}

\begin{document}

\maketitle

\input{Sections/0_Abstract}

\input{Sections/1_Introduction}

\input{Sections/2_Software_Evolution}

\input{Sections/3_Hardware_Evolution}

\input{Sections/4_Metrics}

\input{Sections/5_Applications}

\input{Sections/6_Conclusion}

\input{Sections/7_Additional_Parts}

\clearpage

\bibliographystyle{naturemag}
\bibliography{Sections/8_References}

\clearpage

\setcounter{page}{1}

\title{Vision-centric generative AI models:\\
 A software-hardware perspective\\\textbf{Supplementary Information}}

\author{
    Eleni Tselepi,
    Cristian Sestito,\\
    Shady Agwa, 
    Themis Prodromakis
}

\date{
    \small
    Centre for Electronics Frontiers, Institute for Integrated Micro and Nano Systems, \\ School of Engineering, The University of Edinburgh, UK
}

\maketitle

\renewcommand{\thefigure}{S\arabic{figure}}
\setcounter{figure}{0} 

\renewcommand{\thetable}{S\arabic{table}}
\setcounter{table}{0} 

\input{Tables/Table_parameters}

\input{Tables/Table_figure_3b}

\input{Tables/Table_section_4}

\end{document}

%% file: Sections/0_Abstract.tex
\begin{abstract}

Vision generative artificial intelligence (AI) has emerged as one of the most rapidly advancing areas of deep learning. The explosion of multimodal models has made them widely associated with text-to-image applications running on large datacentres. However, vision generative models are equally needed in applications that operate under strict hardware constraints at the edge, including autonomous vehicles, agricultural sensors, and mobile devices. In this Perspective, we argue that progress in vision generative AI has been driven by output quality, with hardware evolving reactively to accommodate growing model demands. We quantify the parameter cost and energy efficiency of these models across a range of accelerator platforms, and map four generative model families against seven real-world application domains. Finally, we advocate a software-hardware co-design approach, where deployment constraints are considered from the start of the design process, ensuring that the \textit{``right model''} runs on the \textit{``right hardware''} to serve the \textit{``right application''}, making generative AI deployment sustainable and accessible across a much broader range of platforms.

\end{abstract}

%% file: Sections/1_Introduction.tex
\section*{Introduction}

Vision generative artificial intelligence (AI) refers to the area of deep learning dedicated to creating and enhancing visual content. Applications include text-to-image generation on large datacenters\cite{dalle,stable_diffusion_3}, image reconstruction and super-resolution on edge cameras and mobile devices\cite{image_reconstruction_mobile}, synthetic image generation for medical diagnosis\cite{medical,medical_2}, crop disease detection in agriculture\cite{agriculture}, scene synthesis for autonomous driving perception\cite{Autonomous_driving}, and real-time content generation for augmented reality and virtual reality (AR/VR) headsets\cite{ar/vr}. Throughout the past decade, vision generative AI has scaled in pursuit of better output quality and greater visual realism, with parameter counts, memory requirements, and inference times increasing at each transition\cite{var}. Early models were compact, requiring modest computational resources. As the field progressed, however, each transition brought greater demands. Nowadays, the models that achieve the best benchmark results require substantial memory and inference time (more than 10 GB and several seconds per image\cite{metaverse}), making them impractical on platforms where size, weight, power, and cost (SWaP-C) are tightly constrained, precisely the platforms on which many real-world applications must run.

In parallel, the hardware platforms supporting these models have evolved to accommodate the growing demands of successive model generations. Early neural network workloads ran on central processing units (CPUs), but the growing parallelism requirements of deep learning drove a shift to graphics processing units (GPUs), which offered the memory bandwidth and throughput needed for large-scale matrix operations\cite{alexnet}. Custom application-specific integrated circuits (ASICs)\cite{asic_kim,guo} and Tensor Processing Units (TPUs)\cite{tpu_diff} introduced specialised compute architectures optimised for the matrix-heavy workloads that dominate neural network inference, offering better energy efficiency than general-purpose GPUs\cite{tpu_better_gpu}. Near-memory and neuromorphic architectures emerged more recently\cite{prime,loihi} to address the bandwidth bottlenecks that large models impose, bringing computation closer to where data is stored. However, all these changes happened only after a model was established. New hardware platforms were designed in response to the models that already existed. This resulted into a growing gap between the models the field has converged on and the platforms on which generative AI is increasingly expected to operate.

In this Perspective, we trace the evolution of vision generative models since variational autoencoders (VAEs) introduced the first widely adopted deep generative framework for image synthesis and showcase this alongside the hardware evolution. We build a comprehensive landscape for the vision-centric generative AI model to incorporate the trade-off between computational complexity, targeting higher output quality, and hardware cost, aiming at higher energy efficiency. Consequently, we present a roadmap that categorizes vision-centric generative AI into four families, which cover seven real-world application domains, thereby highlighting a concept of \textit{``right model for the right application''}. Finally, we outline three possible trajectories for the future of vision generative AI, and advocate that software-hardware co-design is essential to make generative AI deployable across a much broader range of platforms and applications, in a sustainable fashion.

%% file: Sections/2_Software_Evolution.tex
\section*{Evolution of vision generative AI models}

\input{Figures/Main/Figure_1}

The development of vision generative AI has progressed through four dominant model families: VAEs, generative adversarial networks (GANs), diffusion models, and autoregressive transformer generators (Fig. \ref{Figure_1}a). Each transition was motivated primarily by improvements in output quality and visual realism. The field moved from relatively compact single-pass latent models to adversarial systems with more complex training dynamics, then to iterative denoising, and finally to attention-centric backbones, with rising demands on compute, memory, and data movement, with much less attention paid to the hardware implications of these shifts.

VAEs\cite{vae}, introduced in 2013, provided one of the first widely adopted deep generative frameworks for image synthesis. They learn to compress an image into a compact latent representation and then reconstruct it, allowing new images to be generated by sampling from that learned space (Fig. \ref{Figure_1}b). This process requires only a single forward pass through the network, making VAEs computationally efficient and useful for early tasks such as compression, denoising, and feature extraction. Their main limitation, however, is perceptual quality. As these models are trained to minimise the average difference between the original and reconstructed image at the pixel level, generated samples often lack fine detail and appear overly smooth. This limitation created the opportunity for a shift, one year later. 

GANs\cite{gan}, introduced in 2014, replaced reconstruction objectives with an adversarial framework in which a generator and discriminator are trained in competition (Fig. \ref{Figure_1}c). This shift allowed the generator to produce images that matched the natural images more closely and substantially improved sharpness relative to VAEs. Although GANs originally used fully connected layers, in which every neuron processes the entire input without spatial awareness,  the introduction of convolutional layers in Deep Convolutional Generative Adversarial Networks (DCGAN)\cite{dcgan} in 2015 made adversarial image generation far more effective. Unlike fully connected layers, convolutional layers operate on local image regions, learning to detect spatially structured features such as edges, textures, and shapes.  Over the following years, GANs became the dominant paradigm for image to image translation, super resolution, data augmentation, face generation, and style transfer. The StyleGAN\cite{stylegan} family, introduced in 2018, pushed GAN-based synthesis to state-of-the-art levels as measured by the Fréchet Inception Distance (FID), a widely used metric that quantifies the similarity between the generated and real images. Importantly, GANs retained a major systems advantage: once trained, they generate images in a single forward pass, making them highly attractive for low-latency deployment. Their weaknesses is concentrated in training rather than deployment. Adversarial optimization is unstable, the generator and discriminator have to improve at a similar pace, and mode collapse can reduce sample diversity.

Diffusion models, particularly Denoising Diffusion Probabilistic Models (DDPMs)\cite{ddpm} introduced in 2020, offered a different solution to the limitations of GANs. Rather than learning through adversarial competition, they are trained to reverse a gradual noising process by repeatedly predicting and removing noise (Fig. \ref{Figure_1}d). This objective proved more stable and scalable than adversarial training, and diffusion models soon surpassed GANs on quality and benchmarks diversity\cite{diffusion_beat_gan}, especially for complex image distributions. The system cost, however, was substantial. Unlike VAEs and GANs, which generate in a single pass, diffusion models require many sequential denoising steps at inference time, often involving tens to hundreds of network evaluations per image. This introduced a step change in latency and memory traffic.

In 2021,  Contrastive Language-Image Pre-training (CLIP)\cite{clip} helped establish a practical bridge between language and vision, enabling semantic control through text and marking an early step towards multimodal generation. In the same year, early systems such as DALL·E\cite{dalle} explored autoregressive transformers for text to image generation, producing images token by token (Fig. \ref{Figure_1}e). In contrast to convolutional layers, which detect local spatial features, transformers employ self-attention to model relationships across the entire image simultaneously, enabling the capture of long range dependencies. These models demonstrated that transformers could serve as generative backbones for vision tasks, but their autoregressive nature meant that generating a single image required a forward pass for each token in the output sequence, making them computationally expensive to train and slow at inference. By 2022, the field shifted back towards diffusion-based pipelines, which offered high image quality at lower inference cost\cite{dalle2}. Early diffusion models operated directly in pixel space, so every denoising step processed the full-resolution image, making generation extremely expensive. Latent Diffusion Models\cite{latent_diffusion} addressed this by first compressing the image into a lower-dimensional latent representation using a VAE, and then performing the diffusion process in that latent space. This reduced computational cost substantially while preserving high visual quality.

By 2023, transformer architectures re-emerged through diffusion transformers (DiTs)\cite{dit}. In these models, the conventional convolutional U-Net backbone was replaced by a transformer operating over tokenized latent image patches during the denoising process (Fig. \ref{Figure_1}d).  By 2024, two distinct trajectories had emerged. One was the diffusion transformer path, exemplified by Stable Diffusion 3\cite{stable_diffusion_3}, in which diffusion remained the generative process while the backbone became increasingly transformer-based. The other was a renewed autoregressive path, represented by visual autoregressive (VAR) models\cite{var}, which revisited token-based image generation and reported competitive results on ImageNet at 256×256 resolution while offering faster inference than DiT baselines.

%% file: Figures/Main/Figure_1.tex
\begin{figure}[t!]
\includegraphics[width=\textwidth]{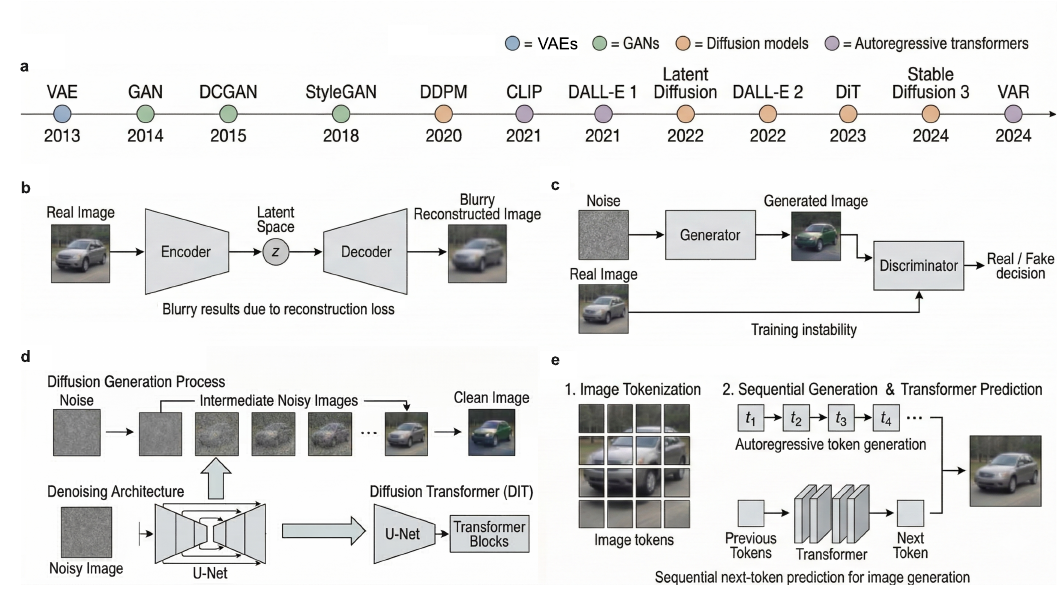}
\centering
\caption{
\textbf{Evolution and core architectures of vision generative models}. 
 \textbf{a,} Timeline of landmark generative models from 2013 to 2024, with each model represented as a point coloured by architectural family: Variational autoencoders (VAE; blue), generative adversarial networks (GAN; green), diffusion models (orange), and autoregressive transformer models (purple). \textbf{b,} VAE architecture, in which an encoder maps the input image to a probabilistic latent space and a decoder reconstructs the image from a sampled latent vector. \textbf{c,} GAN architecture, comprising a generator that synthesises images from random noise and a discriminator trained to distinguish generated from real images through an adversarial objective. \textbf{d,} Diffusion model architecture, in which an image is progressively corrupted by Gaussian noise during a forward process and a neural network learns to reverse this process iteratively to generate new samples. The arrow indicates the architectural evolution toward the diffusion transformer (DiT), in which the conventional U-Net backbone is replaced by a transformer  \textbf{e,} Autoregressive transformer architecture, in which images are represented as sequences of discrete tokens and generated autoregressively using a transformer trained with a next-token prediction objective. 
}
\label{Figure_1}
\end{figure}

%% file: Sections/3_Hardware_Evolution.tex
\section*{Evolution of AI hardware accelerators}

\input{Figures/Main/Figure_2}

It is interesting to note that the hardware platforms supporting vision generative AI models evolved independently from each other. As documented in prior work\cite{the_hardware_lottery}, hardware and software have historically been treated as separate design choices, with hardware investment viable only when the target use case is expected to persist for more than three years. This pattern is clearly visible in the history of hardware AI platforms: accelerators were adapted only after new model classes had already become established, and the constraints of available hardware were rarely consulted before architectural decisions were made (Fig. \ref{Figure_2}).

When VAEs were introduced in 2013, deep learning training had already started to shift from CPUs towards GPUs. Their high arithmetic throughput and memory bandwidth made them far better suited than CPUs to the convolutions and matrix multiplications that dominated deep networks. This is illustrated by the contrast between a 2012 system that required 16,000 CPU cores to classify images and a model published the following year that achieved the same task using just 2 CPU cores and 4 GPUs\cite{the_hardware_lottery}. However, this shift did not occur because GPUs were designed for generative modelling, but because early generative models were small enough to fit within accelerator memory, and their feed-forward structure mapped naturally onto massively parallel devices such as NVIDIA Kepler, which delivered teraflop-scale performance.

In parallel, alternative directions also emerged. IBM TrueNorth\cite{truenorth}, introduced in 2014, represented a neuromorphic path built around extreme energy efficiency rather than raw computational throughput, integrating 1 million neurons while consuming only 70 mW. When tested on real-world problems such as multi-object detection and classification, the chip consumed just 63 mW on a 30-frame-per-second video stream\cite{truenorth}. Microsoft Catapult\cite{catapult}, deployed in 2015, demonstrated that FPGAs could be used at datacentre scale, offering reconfigurable acceleration with lower latency and improved performance per watt, primarily to accelerate search ranking workloads\cite{catapult}. A more pronounced move towards domain-specific acceleration followed. Google's first TPU\cite{tpu1}, in 2015, was a systolic-array inference accelerator designed explicitly around the matrix multiply-accumulate operations, supporting production applications including search ranking, image processing, and natural language translation. Its architecture prioritised energy efficiency rather than floating-point throughput, reflecting a design philosophy distinct from that of GPUs.

Around the same time, research-stage compute-in-memory (CIM) architectures such as ISAAC\cite{isaac} and PRIME\cite{prime}, both published in 2016, targeted a different emerging bottleneck: the memory wall. As model parameter counts grew, the cost of moving weights between off-chip dynamic random access memory (DRAM) and on-chip compute units came to dominate both latency and energy. ISAAC demonstrated this on convolutional neural network inference workloads\cite{isaac}, and PRIME on neural network inference within resistive memory arrays\cite{prime}, both proposing to perform multiply-accumulate operations directly in memory to reduce off-chip data movement.

The progressive scaling of GANs exposed the limits of general purpose GPU compute. NVIDIA Volta, introduced in 2017, added Tensor Cores, dedicated units for small matrix multiply-accumulate operations, and delivered 125 TFLOPS of FP16 throughput, a step change over the Kepler baseline. Volta made large-scale GAN training, including later architectures such as StyleGAN, far more tractable and helped establish mixed-precision training as a standard. Released in the same year, Intel Loihi\cite{loihi} offered a contrasting research direction through neuromorphic computation with on-chip learning and very low power operation. The Cerebras CS-1, introduced in 2019\cite{cerebras}, represented a categorical departure from conventional accelerator design: a wafer-scale processor containing 400,000 AI-optimised cores and 18 GB of on-chip static random access memory (SRAM), greatly reducing dependence on off-chip communication for models that fit within its on-chip capacity. Around the same period, Groq\cite{groq} introduced a tensor-streaming architecture optimized for low-latency inference.

The arrival of diffusion models in 2020 fundamentally changed the computational character of image generation, replacing single forward passes with iterative denoising processes requiring tens to hundreds of sequential network evaluations per sample. Transformer-based generators compounded this by increasing dependence on self-attention, with its associated pressure on memory capacity, bandwidth, and data movement. Existing GPUs and ASICs could still run these models, but increasingly through scale rather than efficiency, relying on larger clusters and high-bandwidth memory. NVIDIA Ampere arrived in 2020, the same year as DDPMs, delivering major gains in tensor throughput and memory bandwidth. The A100 GPU was not, however, designed in response to diffusion models. Nonetheless, diffusion models benefited from its capabilities. A further scaling step came in 2021 with Google TPU v4\cite{tpu4}, which extended the TPU line into systems built for training increasingly large transformer and multimodal models. Intel Loihi 2 advanced the neuromorphic lineage with improved on-chip learning and support for sparse inference workloads, while Samsung HBM-PIM embedded programmable processing elements within the high-bandwidth memory stack, reducing the cost of moving data to a separate compute die. By 2022, hardware had begun adapting more explicitly to transformer workloads. NVIDIA Hopper introduced the Transformer Engine and low-precision support tailored to attention heavy computation, marking a shift from general neural acceleration towards direct support for the model class that was dominant. Alternative efficiency-driven directions continued in parallel. IBM NorthPole\cite{northpole} demonstrated near-memory computation at scale, while Intel Hala Point extended the neuromorphic line to 1.15 billion neurons across 1,152 Loihi 2 chips, making it the largest neuromorphic system to date.  At datacentre scale, NVIDIA Blackwell represented full industry commitment to transformer-class workloads, including Diffusion Transformers (DiTs) and other large generative models.

This chronology shows that compute substrates for vision generative models evolved independently, not following a co-design approach. The accelerator milestones surveyed here do not map onto the four generative model transitions in any consistent way. GPUs supported early VAEs and GANs because their dense arithmetic happened to fit existing accelerators. FPGAs, TPUs, neuromorphic chips, wafer-scale processors, Groq-style tensor-streaming designs, and CIM architectures then emerged as successive responses to rising demands for efficiency, specialization, and reduced data movement. In other words, hardware did not guide the architectural evolution of vision generative AI; it repeatedly absorbed the consequences of choices made elsewhere. This misalignment is not incidental: hardware development cycles of two to three years are structurally incompatible with the pace at which generative model architectures evolve\cite{beyond_moores_law}, making reactive adaptation an inadequate response to the field's current trajectory.

%% file: Figures/Main/Figure_2.tex
\begin{figure}[t]
\includegraphics[width=\textwidth]{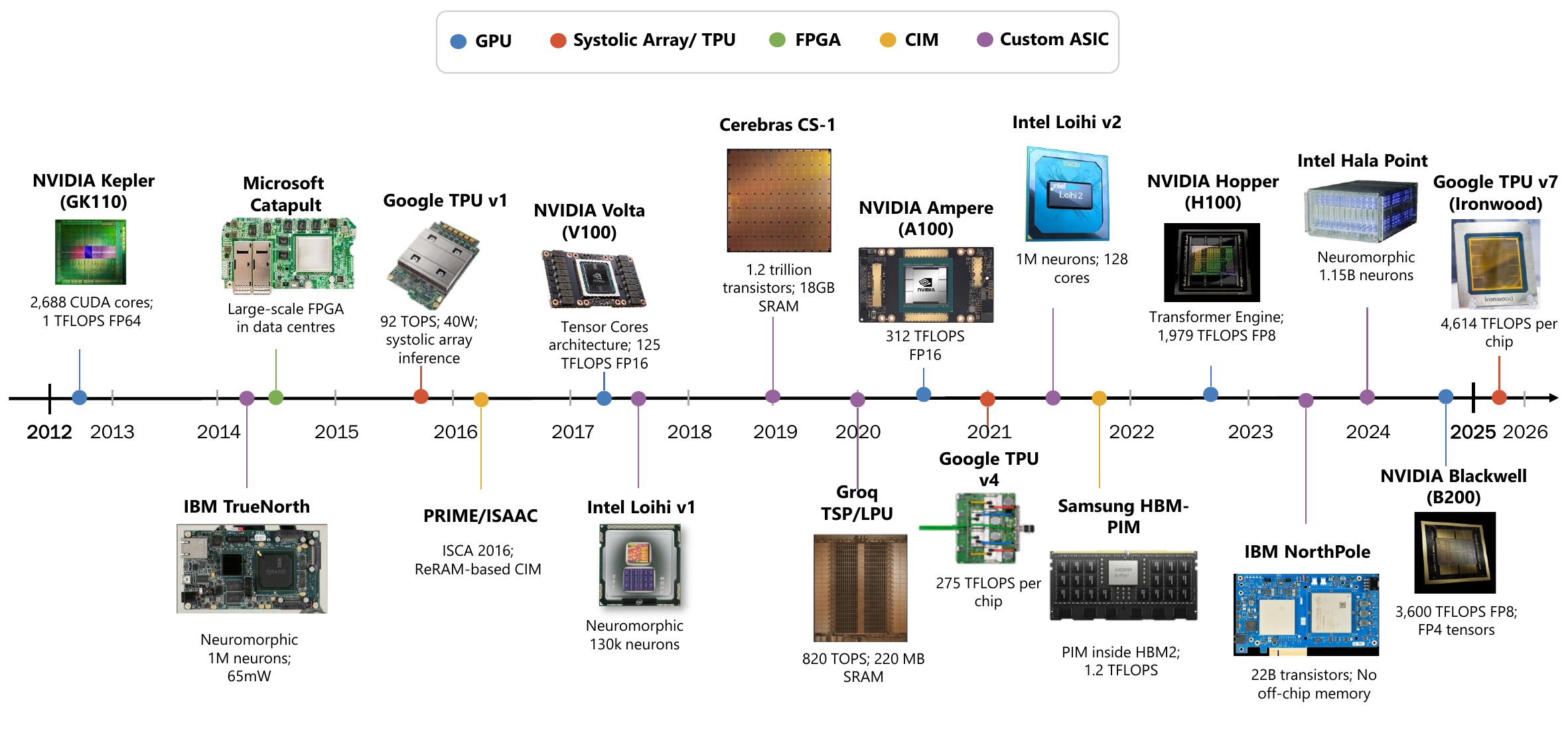}
\centering
\caption{
\textbf{Evolution of AI hardware accelerators.} 
Timeline of key hardware milestones from 2012 to 2025, each annotated with its key architectural contribution, coloured by architectural family: GPUs (blue), systolic arrays and TPUs (red), FPGAs (green), near-memory and compute-in-memory architectures (CIM; orange), and custom ASICs including neuromorphic processors (purple). The timeline reveals a reactive pattern: each wave of hardware innovation followed, rather than preceded, the model demands that motivated it.
 }
\label{Figure_2}
\end{figure}

%% file: Sections/4_Metrics.tex
\section*{The cost of unconstrained quality optimization}

\input{Figures/Main/Figure_3}

Throughout the architectural transitions from VAE to GAN, from GAN to diffusion, and from diffusion to diffusion transformers and autoregressive models, output quality and visual realism improved consistently. However, the hardware cost of that progress received comparatively little attention across these transitions. To address this, we examine model parameter counts against FID, a widely used accuracy metric that evaluates output quality by comparing the statistical distance between generated and real image feature distributions, where lower values indicate that generated images are statistically closer to real ones (Fig. \ref{Figure_3}a and Supplementary Table \ref{Table_S1}). To ensure a fair comparison across model families, FID scores are reported across three benchmark datasets: CIFAR-10, ImageNet 128×128, and ImageNet 256×256. Examining the four model families reveals that quality gains and parameter cost do not scale proportionally: more parameters do not guarantee better output quality, and in some cases, larger models achieve worse FID scores than smaller ones, suggesting that strong results can be achieved with fewer parameters. 

GAN models establish the most parameter-efficient frontier in this comparison, occupying the lowest-parameter region across all three datasets while maintaining competitive FID scores. On CIFAR-10 and ImageNet 128×128, GAN models achieve the best FID results across all three model families. The most compact GAN model in the CIFAR-10 dataset achieves a competitive FID of 2.64 at just 9.4 million parameters, a score comparable to diffusion models that require three to twelve times more parameters to reach similar results\cite{studiogan}. At these scales, parameter efficiency and output quality can be achieved simultaneously.

As image resolution increases, diffusion and autoregressive models become more competitive. On ImageNet 256×256, the best performing GAN model achieves an FID of 2.32 at 166 million parameters\cite{studiogan}. At a comparable parameter budget, the best-performing autoregressive model achieves an FID of 3.48 at 208 million parameters, performing worse than the leading GAN despite using more parameters\cite {autoregressive_survey}. The best diffusion model improves on the GAN frontier by just 0.22 FID points, yet requires 3 billion parameters to do so\cite{var}, 18 times more than the leading GAN. Among autoregressive models, the best performing model achieves optimum performance in the entire dataset at an FID of 1.55 using 943 million parameters, an improvement of 0.77 FID points over the GAN frontier at nearly six times the model complexity\cite{autoregressive_survey}.  Across all three families, GAN models typically require 100 to 200 million parameters, while diffusion models require 400 million to 700 million, and transformer models require 1 billion to 4 billion parameters for comparable or marginally better results (Fig. \ref{Figure_3}a), reflecting how complexity has grown substantially with each model transition.

Parameter count alone does not, however, determine deployment feasibility. The energy efficiency of a model-platform pairing is equally consequential, and when evaluated across different hardware platforms, the energy efficiency of the same generative model can differ by several orders of magnitude (Fig. \ref{Figure_3}b and Supplementary Table \ref{Table_S2}). Custom ASIC designs achieve the highest energy efficiency across all model families, operating mostly in the range of 10 TOPS/W to 100 TOPS/W. GAN models on ASICs represent the most efficient combination, reaching 135.1 TOPS/W\cite{ganpu}. Diffusion models on the same hardware demonstrate that specialised silicon can partially compensate for the higher computational cost of iterative generation, achieving close to 100 TOPS/W\cite{guo}, while transformer models on ASICs reach approximately 40 TOPS/W\cite{cimformer}. Across all three model families, ASIC deployment delivers substantial efficiency gains, suggesting that hardware specialisation is a viable path toward efficient generative inference. TPU implementations achieve energy efficiency comparable to ASIC deployments, operating at around 10 TOPS/W\cite{tpu_diff}, reflecting the advantages of systolic array design for matrix-heavy workloads. GPU platforms\cite{mlenergy} deliver the highest absolute throughput but show lower energy efficiency than ASICs and TPUs, as their high power consumption offsets their throughput advantage. FPGA implementations are three orders of magnitude less energy efficient than their ASIC counterparts, operating between 10 GOPS/W and 100 GOPS/W\cite{fpga_kim,sda,shao} across all model families, and deliver the lowest absolute throughput in the dataset. The ordering across platforms is consistent regardless of which generative architecture is deployed, with ASIC designs achieving the highest energy efficiency across all model families, reinforcing the conclusion that hardware choice is as consequential as model choice for deployment feasibility.

%% file: Figures/Main/Figure_3.tex
\begin{figure}[t]
\includegraphics[width=\textwidth]{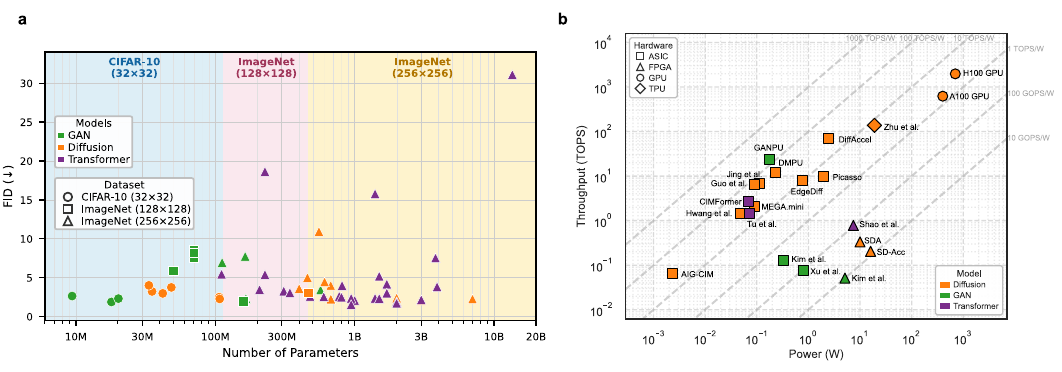}
\centering
\caption{
\textbf{Parameter efficiency and computational performance of generative vision models.} 
\textbf{a,} Fréchet inception distance (FID; lower values indicate closer distributional alignment with real data) versus number of parameters for generative models across three benchmark datasets: CIFAR-10 (blue background, circle markers), ImageNet 128$\times$128 (pink background, square markers), and ImageNet $256\times$256 (yellow background, triangle markers). Generative adversarial networks (GANs), diffusion, and autoregressive (transformer) models are shown in green, orange, and purple, respectively. \textbf{b,} Computational throughput (TOPS) versus power consumption (watts) for representative model and hardware platform pairs. Diagonal dotted lines represent energy efficiency levels of 10 GOPS/W, 100 GOPS/W, 1 TOPS/W, 10 TOPS/W, 100 TOPS/W, and 1000 TOPS/W. Architecture families follow the same colour encoding as \textbf{a}. Hardware platforms are distinguished by marker shape: circles denote GPUs, triangles denote FPGAs,  squares denote ASICs, and diamonds TPUs. 
}
\label{Figure_3}
\end{figure} 

%% file: Sections/5_Applications.tex
\section*{Matching generative models to application domains}

\input{Figures/Main/Figure_4}

The parameter and energy costs documented here raise the question of whether those costs are justified by the requirements of the applications in which generative models are now expected to operate. To investigate this further, we considered seven real-world domains, selected to cover the three constraints that govern deployment feasibility: output quality, inference latency, and memory capacity. From the application point of view, these domains span cloud content generation, medical imaging, autonomous vehicles, AR/VR, agricultural monitoring, mobile and edge devices, and aerial and remote sensing. Each application sets its own limits across these categories, and the architectures that currently dominate the literature are well-matched to a small minority of them (Fig. \ref{Figure_4}a and Supplementary Table \ref{Table_S3}).

For a number of applications, output quality is the dominant constraint, either because hardware resources are unconstrained and the goal is maximum visual realism, or because the application demands a specific standard of fidelity that cannot be compromised. Cloud-based content generation is the domain where there are no limitations. Power budgets are unconstrained, memory is abundant, and output quality is the primary concern (Fig. \ref{Figure_4}b). For this reason, diffusion and transformer architectures are well-suited to these platforms, as their high memory and compute requirements can be fully accommodated \cite{var,ddpm}. Medical imaging belongs here too, though for different reasons. Generated images must preserve sufficient quality for diagnostic or training use, accurately capturing pathological features and anatomical structures\cite{medical}. GAN models dominated medical image synthesis for much of the past decade, but diffusion and transformer models have since demonstrated stronger performance on tasks requiring fine structural detail\cite{diffuion_better_gan_medical}. Clinical workstations and hospital servers provide GPU-class compute with no significant hardware restrictions. These two domains are where the field's benchmark-driven optimization aligns with deployment reality, but they represent a small area of where generative models are now being asked to operate.

Latency is the binding constraint for a second and larger group of applications. Autonomous vehicles require real-time inference with hard timing budgets of approximately 150ms\cite{automotive}. These platforms carry larger onboard memory than edge devices, which means the feasibility boundary is defined by inference time rather than model size, making GAN models and certain lightweight diffusion architectures the options at this latency threshold. AR/VR applications are more demanding still, with inference times below 20ms needed to maintain high display frame rates\cite{ar/vr, metaverse}. AR/VR applications also place genuine demands on output quality, since users observe the generated content directly, making GAN models well-positioned across both contexts. Agricultural deployment introduces a related timing constraint at a more moderate scale, with generation expected to complete within approximately 500ms\cite{agriculture}. Agricultural platforms also carry limited memory, which compounds the timing constraint and makes GAN and VAE-based architectures the most practical options in this setting (Fig. \ref{Figure_4}c).

Memory capacity defines the constraint for the remaining domains. Mobile and edge devices, including aerial and remote sensing platforms, typically offer between 4 and 16 GB\cite{mobile} of DRAM capacities (Fig. \ref{Figure_4}d). As expected, diffusion and transformer models are poorly matched to this tier. Their parameter counts translate into memory footprints that exceed what these platforms can accommodate, and their inference behavior under memory pressure is difficult to manage in deployment. GAN and VAE-based architectures remain the practical options for hardware in this class. 

Taken together, the seven domains show that the two model families dominating the field's benchmark rankings, diffusion models and transformer-based architectures, are well-matched to at most two deployment contexts. For the remaining five, the perceptual sufficiency criterion is met by architectures that the field has largely abandoned. GAN models are the only family that appears as an option across all seven domains, a fact that current benchmarks do not capture.

%% file: Figures/Main/Figure_4.tex
\begin{figure}[t]
\includegraphics[width=\textwidth]{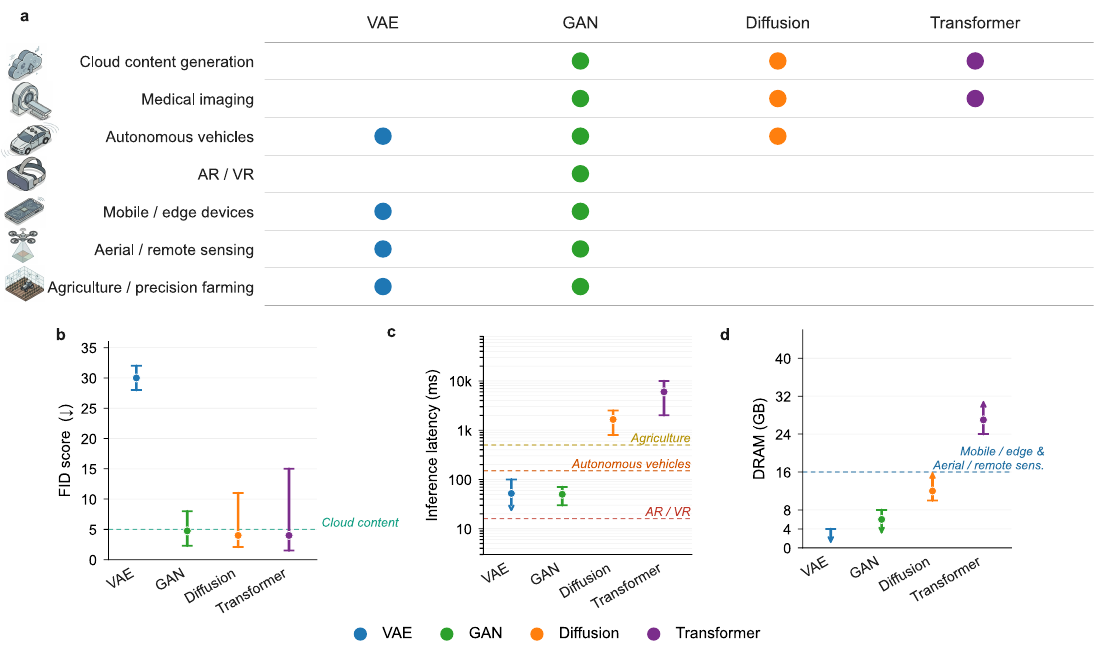}
\centering
\caption{
\textbf{The right model for the right application.} 
\textbf{a,} Compatibility matrix mapping the four generative model families, variational autoencoders (VAEs; blue), generative adversarial networks (GANs; green), diffusion models (orange), and autoregressive transformer models (purple), against seven application domains. Filled circles indicate architectural feasibility within the deployment constraints of each domain.
\textbf{b,} Fréchet inception distance (FID) score upper limits for quality-constrained application domains, overlaid with the FID ranges of the four model families. Applications whose feasibility is primarily determined by output quality are shown.  
\textbf{c,} Inference latency upper limits for latency-constrained application domains, overlaid with the latency ranges of the four model families. Applications whose feasibility is primarily determined by inference time are shown.
\textbf{d,} Memory footprint upper limits for memory-constrained application domains, overlaid with the memory ranges of the four model families. Applications whose feasibility is primarily determined by memory capacity are shown. Together, panels b, c, and d show where each model family falls relative to the maximum acceptable value for each application.
}
\label{Figure_4}
\end{figure}

%% file: Sections/6_Conclusion.tex
\section*{Outlook}

\input{Figures/Main/Figure_5}

The analysis presented here reveals a consistent pattern: as generative AI models progressed through the four architectural families, output quality and visual realism improved, while parameter counts, memory requirements, and inference times grew at each transition. In parallel, hardware accommodated these models, but only after a new model generation had already been established, creating a gap between the generative AI models and the platforms on which they are expected to operate across real-world applications. Understanding how generative AI will progress requires looking beyond model performance and toward the broader consequences for the infrastructure that supports it and the societies that depend on it. To that end, we consider three hypothetical trajectories, each representing a different choice for where efficiency gains could be directed, and each producing a fundamentally different outcome (Fig. \ref{Figure_5}).

The first trajectory describes the present, where models progress independently of the underlying hardware (Fig. \ref{Figure_5}a). They become more capable and more realistic, but this requires substantially larger datacenters and growing energy consumption. Global datacenter electricity consumption is projected to double by 2030, from approximately 1.5\%\cite{energy_consumption} of global electricity consumption in 2024. Leading technology companies have begun investing directly in nuclear energy to secure the reliable, carbon-free power that AI datacentres require. Microsoft restarted the Three Mile Island nuclear plant in Pennsylvania under a 20-year power purchase agreement\cite{microsoft_nuclear}. Google signed their first nuclear energy deal with Kairos Power for small modular reactors to come online in the early 2030s\cite{google_nuclear}. Amazon committed more than 20 billion dollars to nuclear-powered AI infrastructure, and Meta secured agreements for up to 6.6 GW of nuclear capacity\cite{amazon_nuclear,meta_nuclear}. As more applications move to the cloud, societies become increasingly dependent on large centralised datacentre infrastructure, with the energy consequences that entails.

The second trajectory employs the software-hardware co-design at its core (Fig. \ref{Figure_5}b). Rather than focusing on making models more capable, we maintain both the models and the hardware we have today and rethink how they work together, matching the \textit{``right model''} to the \textit{``right hardware''} for the \textit{``right application''}. This approach is already gaining momentum from both a commercial and a policy perspective. The edge AI hardware market is projected to grow from 26 billion dollars in 2025 to nearly 59 billion dollars by 2030\cite{edge_ai_market}, driven by demand for real-time and low-latency inference. From a policy perspective, the UK Royal Academy of Engineering, has identified AI efficiency as a foundational priority, calling on governments to set environmental requirements for AI systems and datacentres and to promote sustainable AI deployment\cite{royal_engineering}. Matching generative models to the hardware platforms best suited to their target application is a concrete expression of this principle: it reduces the energy and infrastructure cost of deployment without requiring new models or new hardware, and it unlocks a broader range of applications to run on edge devices, reducing but not eliminating dependence on remote datacentres.

The third trajectory points to the future (Fig. \ref{Figure_5}c). It presents a scenario in which fundamental disruption at both the model and hardware level is required. New foundational models must be redesigned and beyond-CMOS technologies, including neuromorphic computing architectures\cite{memristors} must work in concert to make generative vision AI run locally, without dependence on large datacentres. The commercial momentum behind these technologies is accelerating, with memristive and neuromorphic architectures moving from research prototypes toward embedded commercial products, and commercial in-memory computing applications expected from 2030 onwards\cite{Lanza_memristors}. Designing models and hardware together will enable generative AI to become ambient, present in the full range of applications that need it, and accessible to everyone.

%% file: Figures/Main/Figure_5.tex
\begin{figure}[t]
\includegraphics[width=\textwidth]{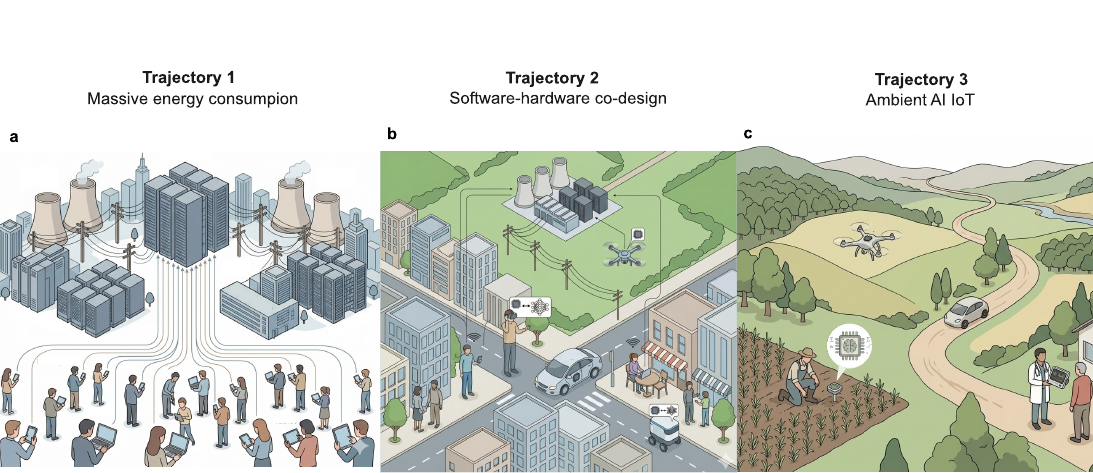}
\centering
\caption{
\textbf{Three possible trajectories for the deployment of efficient generative AI.} 
\textbf{a,} The first trajectory describes the present moment, in which models improve in accuracy alongside growing datacenters infrastructure and energy consumption.
\textbf{b,} The second trajectory represents the software-hardware co-design approach advocated in this Perspective, in which models are matched to the right hardware to serve the right application. Some applications run locally on edge devices, but dependence on remote datacenters persists.
\textbf{c,} The third trajectory describes the ambient future. Post-CMOS computing architectures and algorithmic reconsiderations at the foundational AI level enable applications to run entirely on edge devices effectively, eliminating the need for large datacenters and making intelligence accessible everywhere.
}
\label{Figure_5}
\end{figure}

%% file: Sections/7_Additional_Parts.tex
\section*{Acknowledgments}

This work was supported by the Engineering and Physical Sciences Research Council (EPSRC) AI Hub for Productive Research and Innovation in eLectronics (APRIL) under Grant No. EP/Y029763/1, and by the Royal Academy of Engineering (RAEng) Chair in Emerging Technologies under Grant No. CiET1819/2/93. Figures \ref{Figure_1} and \ref{Figure_5} were created using Google Gemini and subsequently refined by the authors.

\section*{Competing interests}

The authors declare no conflict of interest.

%% file: Tables/Table_parameters.tex
\clearpage
\begin{tabularx}{\textwidth}{|l|X|c|c|}
\caption{FID scores and parameter counts for generative models on CIFAR-10, ImageNet 128$\times$128, and ImageNet 256$\times$256.}
\label{Table_S1} \\

\hline
\textbf{Dataset} & \textbf{Model} & \textbf{Params} & \textbf{FID $\downarrow$} \\ \hline
\endfirsthead

\hline
\textbf{Dataset} & \textbf{Model} & \textbf{Params} & \textbf{FID $\downarrow$} \\ \hline
\endhead

\hline \multicolumn{4}{r|}{\textit{Continued on next page}} \\
\endfoot

\hline
\endlastfoot

CIFAR-10
    & \multicolumn{3}{l|}{\textit{Diffusion}} \\ \cline{2-4}
    & DDPM\cite{studiogan}            & 35.2M   & 3.23 \\
    & DDPM++\cite{studiogan}          & 106.6M  & 2.49 \\
    & NCSN++\cite{studiogan}          & 107.6M  & 2.27 \\
    & DDGAN     \cite{tacking}      & 48.4M   & 3.75 \\
    & WDDGAN   \cite{wavelet}       & 33.4M   & 4.01 \\
    & LDDGAN     \cite{latent}     & 42.0M   & 2.98 \\ \cline{2-4}
    & \multicolumn{3}{l|}{\textit{GAN}} \\ \cline{2-4}
    & StyleGAN-XL\cite{studiogan}     & \textbf {18.0M}   & \textbf{1.88} \\
    & StyleGAN2-ADA\cite{studiogan}   & 20.2M   & 2.31 \\
    & ReACGAN+DiffAug\cite{studiogan} & 9.4M    & 2.64 \\ \hline

\shortstack[l]{ImageNet\\128$\times$128}
    & \multicolumn{3}{l|}{\textit{GAN}} \\ \cline{2-4}
    & BigGAN\cite{studiogan}        & 70M     & 7.66 \\
    & BigGAN\cite{studiogan}          & 70M     & 8.54 \\
    & BigGAN-Deep\cite{studiogan}    & 50M     & 5.90 \\
    & ReACGAN\cite{studiogan}         & 70M     & 8.19 \\
    & StyleGAN-XL\cite{studiogan}    & \textbf{159M}    & \textbf{1.94} \\ \cline{2-4}
    & \multicolumn{3}{l|}{\textit{Diffusion}} \\ \cline{2-4}
    & ADM-G\cite{studiogan}           & 464M    & 3.05 \\ \hline

\shortstack[l]{ImageNet\\256$\times$256}
    & \multicolumn{3}{l|}{\textit{Autoregressive}} \\ \cline{2-4}
    & VAR-d16 \cite{var}        & 310M    & 3.30 \\
    & VAR-d20  \cite{var}        & 600M    & 2.57 \\
    & VAR-d24  \cite{var}        & 1B      & 2.09 \\
    & VAR-d30  \cite{var}        & 2B      & 1.92 \\
    & VAR-d30-re  \cite{var}     & 2B      & 1.73 \\
    & VQVAE-2  \cite{var}       & 13.5B   & 31.11 \\
    & VQGAN     \cite{var}       & 227M    & 18.65 \\
    & VQGAN    \cite{var}        & 1.4B    & 15.78 \\
    & VQGAN    \cite{studiogan}       & 1.5B    & 5.20 \\
    & ViTVQ   \cite{var}         & 1.7B    & 4.17 \\
    & ViTVQ-re   \cite{var}     & 1.7B    & 3.04 \\
    & RQ-Transformer \cite{var}   & 3.8B    & 7.55 \\
    & RQ-Transformer \cite{studiogan} & 3.9B    & 3.83 \\
    & MaskGIT   \cite{studiogan}      & 227M    & 5.40 \\
    & LlamaGen-B   \cite{autoregressive_survey}   & 111M    & 5.46 \\
    & LlamaGen-L  \cite{autoregressive_survey}    & 343M    & 3.07 \\
    & LlamaGen-XL  \cite{autoregressive_survey}   & 775M    & 2.62 \\
    & LlamaGen-XXL \cite{autoregressive_survey}   & 1.4B    & 2.34 \\
    & LlamaGen-3B  \cite{autoregressive_survey}   & 3B      & 2.18 \\
    & Open-MAGVIT2-B    \cite{autoregressive_survey} & 343M    & 3.08 \\
    & Open-MAGVIT2-L \cite{autoregressive_survey}  & 804M    & 2.51 \\
    & Open-MAGVIT2-XL \cite{autoregressive_survey} & 1.5B    & 2.33 \\
    & MAR-B   \cite{autoregressive_survey}        & 208M    & 3.48 \\
    & MAR-L \cite{autoregressive_survey}          & 479M    & 2.60 \\
    & MAR-H \cite{autoregressive_survey}          & 943M    & 2.35 \\
    & MAR-H w/ CFG  \cite{autoregressive_survey}  & \textbf{943M}    & \textbf{1.55} \\
    & DART w/ CFG   \cite{autoregressive_survey}  & 812M    & 3.98 \\ \cline{2-4}
    & \multicolumn{3}{l|}{\textit{Diffusion}} \\ \cline{2-4}
    & DiT-XL/2 \cite{var}       & 675M    & 2.27 \\
    & DiT-L/2 \cite{var}          & 458M    & 5.02 \\
    & ADM \cite{var}          & 554M    & 10.94 \\
    & LDM-4-G \cite{var}         & 400M    & 3.60 \\
    & VDM++  \cite{autoregressive_survey}      & 2B      & 2.40 \\
    & L-DiT-3B  \cite{var}       & 3B      & 2.10 \\
    & L-DiT-7B  \cite{var}       & 7B      & 2.28 \\
    & ADM-G-U \cite{studiogan}        & 673M    & 4.01 \\
    & ADM-G  \cite{studiogan}         & 608M    & 4.48 \\ \cline{2-4} 
    & \multicolumn{3}{l|}{\textit{GAN}} \\ \cline{2-4}
    & BigGAN\cite{studiogan}          & 164M    & 7.75 \\
    & BigGAN-Deep\cite{studiogan}     & 112M    & 6.95 \\
    & StyleGAN-XL\cite{studiogan}     & 166M    & 2.32 \\
    & GigaGAN\cite{var}         & 569M    & 3.45 \\ \hline

\end{tabularx}

%% file: Tables/Table_figure_3b.tex
\clearpage
\begin{tabularx}{\textwidth}{|>{\centering\arraybackslash}p{2cm}|>{\centering\arraybackslash}p{3cm}|>{\centering\arraybackslash}p{3cm}|>{\centering\arraybackslash}p{3cm}|>{\centering\arraybackslash}p{4cm}|}
\caption{Hardware accelerators for vision generative AI: throughput, power consumption, and energy efficiency.}
\label{Table_S2} \\
\hline
\textbf{Hardware Type} & \textbf{Model} & \textbf{Throughput (TOPS)} & \textbf{Power (W)} & \textbf{Energy Efficiency (TOPS/W)} \\ \hline
\endfirsthead
\hline
\textbf{Hardware Type} & \textbf{Model} & \textbf{Throughput (TOPS)} & \textbf{Power (W)} & \textbf{Energy Efficiency (TOPS/W)} \\ \hline
\endhead
\hline \multicolumn{5}{r|}{\textit{Continued on next page}} \\
\endfoot
\hline
\endlastfoot
\multirow{4}{*}{GPU}
    & \multicolumn{4}{l|}{\textit{NVIDIA A100}} \\ \cline{2-5}
    & \mbox{Stable Diffusion} \cite{mlenergy}   & 624       & 400   & 1.56 \\ \cline{2-5}
    & \multicolumn{4}{l|}{\textit{NVIDIA H100}} \\ \cline{2-5}
    & \mbox{Stable Diffusion} \cite{mlenergy}   & 1978.9    & 700   & 2.83 \\ \hline
\multirow{2}{*}{TPU}
    & \multicolumn{4}{l|}{\textit{Diffusion}} \\ \cline{2-5}
    & Zhu et al. \cite{tpu_diff}                & 138       & 19    & 7.26 \\ \hline
\multirow{5}{*}{FPGA}
    & \multicolumn{4}{l|}{\textit{GAN}} \\ \cline{2-5}
    & Kim et al. \cite{fpga_kim}                & 0.0512    & 5.1   & 0.01004 \\ \cline{2-5}
    & \multicolumn{4}{l|}{\textit{Diffusion}} \\ \cline{2-5}
    & SDA \cite{sda}                            & 0.3325    & 9.977 & 0.03332 \\
    & SD-Acc \cite{sd-acc}                      & 0.2048    & 15.98 & 0.01281 \\ 
    \cline{2-5}
    & \multicolumn{4}{l|}{\textit{Transformer}} \\ \cline{2-5}
    & Shao et al. \cite{shao}                   & 0.7802    & 7.43  & 0.1051 \\ 
    \hline
\multirow{12}{*}{ASIC}
    & \multicolumn{4}{l|}{\textit{GAN}} \\ \cline{2-5}
    & Xu et al. \cite{xu}                       & 0.0769    & 0.805 & 0.09552 \\
    & Kim et al. \cite{asic_kim}                & 0.1248    & 0.325 & 0.38 \\
    & GANPU \cite{ganpu}                        & 1.08 -- 24.13     & 0.058 -- 0.647 & 1.66 -- 135.10 \\ \cline{2-5}
    & \multicolumn{4}{l|}{\textit{Diffusion}} \\ \cline{2-5}
    & \mbox{MEGA.mini} \cite{megamini}          & 2.048     & 0.088 -- 0.172 & 11.9 -- 23.0 \\
    & AIG-CIM \cite{aig-cim}                    & 0.064     & 0.004 & 14 -- 27 \\
    & Jing et al. \cite{jing}                   & 6.79 -- 9.71  & 0.074 -- 0.211 & 49.74 -- 60.81 \\
    & DMPU \cite{dmpu}                          & 0.47 -- 12.02 & 0.060 -- 0.279 & 1.67 -- 52.01 \\
    & Guo et al. \cite{guo}                     & 4.424 -- 6.636    & 0.008 -- 0.171 & 67.89 -- 74.34 \\
    & DiffAccel \cite{diffaccel}                & 69        & 2.47  & 27.94 \\
    & Picasso \cite{picasso}                    & 9.83      & 1.98  & 4.96 \\ 
    & EdgeDiff \cite{edgediff}                  & 7.8       & 0.78  & 10 \\ 
    & Hwang et al. \cite{hwang}                 & 0.06      & 0.0019 & 31.2 \\ \cline{2-5}
    & \multicolumn{4}{l|}{\textit{Transformer}} \\ \cline{2-5}
    & Tu et al. \cite{tu}                       & 1.48      & 0.027 -- 0.118 & 20.5 \\ 
    & CIMFormer \cite{cimformer}                & 2.66      & 0.0204 -- 0.0973 & 38.9 \\ \hline
\end{tabularx}

%% file: Tables/Table_section_4.tex
\clearpage
\begin{longtable}{|l|l|c|c|c|}
\caption{Model metrics and application deployment requirements 
corresponding to Figures~\ref{Figure_4}a--d. Inference time and DRAM 
ranges for each model family are reported from\cite{metaverse}. 
FID scores are evaluated on the ImageNet 256$\times$256 dataset. 
Inference time is measured for a single 512$\times$512 image on an 
NVIDIA A100 GPU. For each 
application, only the primary binding constraint ($\star$) is reported. 
FID thresholds for cloud content generation represent 
perceptual sufficiency bounds defined by the authors.}
\label{Table_S3} \\

\hline
\textbf{Category} & \textbf{Model / Application} & \textbf{FID ($\downarrow$)} & \textbf{Inference Time ($\downarrow$)} & \textbf{DRAM (GB)} \\ \hline
\endfirsthead

\hline
\textbf{Category} & \textbf{Model / Application} & \textbf{FID ($\downarrow$)} & \textbf{Inference Time ($\downarrow$)} & \textbf{DRAM (GB)} \\ \hline
\endhead

\hline \multicolumn{5}{r|}{\textit{Continued on next page}} \\
\endfoot

\hline
\endlastfoot

\multirow{4}{*}{Model families}
    & VAE         & 28--32  & $<$100 ms              & $<$4            \\
    & GAN         & 2.3--8  & 30--70 ms             & $<$8            \\
    & Diffusion   & 2.1--11 & 0.8--2.5 s        & $\sim$10        \\
    & Transformer & 1.5--15 & 2--10 s   & $>24$  \\ \hline

\multirow{9}{*}{\shortstack[l]{Application\\requirements}}
& \multicolumn{4}{l|}{\textit{Quality-constrained}} \\ \cline{2-5}
    & Cloud content gen.      & $<$5$\star$  & ---  & ---  \\ \cline{2-5}
    & \multicolumn{4}{l|}{\textit{Latency-constrained}} \\ \cline{2-5}
    & AR / VR                 & ---  & $<$16ms$\star$$^{a}$    & ---  \\
    & Autonomous vehicles     & ---  & $<$150 ms$\star$$^{b}$   & ---  \\
    & Agriculture             & ---  & $<$500 ms$\star$$^{c}$   & ---  \\ \cline{2-5}
    & \multicolumn{4}{l|}{\textit{Memory-constrained}} \\ \cline{2-5}
    & Mobile / edge devices   & ---  & ---  & 4--16$\star$$^{d}$  \\
    & Aerial / remote sensing & ---  & ---  & 4--16$\star$$^{d}$   \\ \hline

\end{longtable}

\noindent\small$^{a}$Real-time AR/VR requires latency $<$16\,ms per frame ($>$60\,fps)~\cite{ar/vr, metaverse}.\\
$^{b}$Autonomous vehicle systems  expected reaction times ~\cite{automotive}.\\
$^{c}$Inference time for precision 
spraying~\cite{agriculture}.\\
$^{d}$Device DRAM budget for mobile and edge 
deployment~\cite{mobile}.